\documentclass[11pt]{article}

\usepackage[]{ranlp2021}

\usepackage{times}
\usepackage{latexsym}
\usepackage{amsthm}
\usepackage{amsmath, amssymb}
\usepackage{graphicx}
\newtheorem{theorem}{Theorem}

\aclfinalcopy
\usepackage[T1]{fontenc}

\usepackage[utf8]{inputenc}

\usepackage{microtype}

\title{On the Evolution of Word Order}

\author{Idan Rejwan \\
  Tel Aviv University and AI21 Labs \\
  Tel Aviv, Israel \\
  \texttt{idanrejwan91@gmail.com} \\\And
  Avi Caciularu \\
  Bar-Ilan University \\ Ramat-Gan, Israel \\
  \texttt{avi.c33@gmail.com} \\}

\date{}

\begin{document}
\maketitle
\begin{abstract}
Most natural languages have a predominant or fixed word order. For example in English the word order is usually Subject-Verb-Object. This work attempts to explain this phenomenon as well as other typological findings regarding word order from a functional perspective. In particular, we examine whether fixed word order provides a functional advantage, explaining why these languages are prevalent.
To this end, we consider an evolutionary model of language and demonstrate, both theoretically and using genetic algorithms, that a language with a fixed word order is optimal. We also show that adding information to the sentence, such as case markers and noun-verb distinction, reduces the need for fixed word order, in accordance with the typological findings.
\end{abstract}

\section{Introduction}

Word order is a linguistic concept that refers to how syntactic elements are arranged in a sentence in different languages. Specifically, it deals with the relative position of the three prominent syntactic roles in a sentence: subject (\emph{S}), verb (V), and object (O).
The three syntactic roles can be ordered in six different arrangements: \emph{SVO, SOV, VSO, VOS, OVS, OSV}. Apriori, there is no reason to assume that one order is superior to others. However, typological findings show that most natural languages have one preferred or fixed word order, with the vast majority of them belonging to the first two families: SVO or SOV \cite{comrie1989language}. Interestingly, languages that have developed spontaneously and did not contact other languages also have a distinct preference for particular word order, such as the Al-Sayyid Bedouin Sign Language \cite{sandler2005emergence}. 

Another typological finding is related to case markers, which are morphological indicators (usually prefixes or suffixes) that indicate the syntactic role of a word in a sentence. For example, the word for \emph{the boy} in Arabic is \emph{al-walad-u} if it is the subject and \emph{al-walad-a} if it is the object. Apparently, in languages with a rich system of case markers, such as Arabic, Latin and Sanskrit, the word order is more flexible \cite{sapir1921language, mcfadden2004position}.

These findings raise several research questions. First, why is there a preferred or fixed order in the majority of natural languages? Second, why do most languages have \emph{SVO} or \emph{SOV} as their preferred order? Third, how can we explain the relationship between case markers and flexible word order?

This work addresses the aforementioned questions within an evolutionary perspective \cite{pinker1990natural, nowak2000evolution}. As in the evolution theory of the animal world, if a particular feature appears in many different species, it is assumed to give a functional advantage to the organisms carrying it, thus helping them survive natural selection. Similarly, linguistic phenomena that are common to all or most languages are assumed to provide a functional advantage and facilitate better communication.

Within this perspective, a plethora of works suggested to use Genetic Algorithms (GA), an optimization method inspired by evolutionary principles, in order to investigate typological findings \cite{batali1998computational,kirby1999syntax,levin1995evolution,kirby2001spontaneous}. In this work, we follow this framework and use GA to study the phenomenon of word order, yet not investigated to the best of our knowledge. In particular, we attempt to answer whether a fixed word order gives a functional advantage, which may explain its emergence in most natural languages. We also provide a theoretical analysis to prove that the optimal grammar, with respect to our functional assumptions, is a grammar with fixed word order. Although it would be interesting to explore why a specific word order is preferred, for example SVO or SOV, this work addresses the fundamental question about why languages have fixed word orders at all, rather than specific ones.

\section{Method}

Following \citet{pinker1990natural}, we view language as a form of communication designed to transmit a message between a speaker and a hearer. Thus, if we are to speak of an ``optimal'' language, it is the one that facilitates precise delivery of a message between the speaker and the listener. 

In this view, the hearer observes only the sentence received from the speaker, which does not contain information regarding the relationship between the words and their syntactic roles. 

Intuitively, word order might impact the way the hearer understands the sentence. For example, compare the English sentence ``\emph{we ship a man}'' with ``\emph{we man a ship}''. In this case, as the words ``\emph{ship}'' and ``\emph{man}'' may function as both a noun and a verb, and due to the absence of case markers, the meaning is determined solely by the order of words in the sentence. 

Under these assumptions, we use GA to simulate a language's evolution, starting with no preferred word order, and test whether communication constraints enforce a fixation of particular word order.

\subsection{Algorithm}
The algorithm is initialized with a \emph{lexicon}, a fixed list of 1000 words, where each word is a string of three letters. The words are randomly initialized and are given to all agents. We test both options of one lexicon for all words or two lexicons for nouns and verbs (see elaboration in section \ref{sec:experiments}).

We initialize a population of 100 \emph{grammars}, where each one is a probability distribution over the six possible word orders and is initialized to the uniform probability. Each grammar is represented by a \emph{speaker} and a \emph{hearer} trying to communicate a message over a noisy channel, and their success in communication measures the fitness of the grammar. In other words, for every grammar, the \emph{speaker} produces a sentence based on the grammar, and the \emph{hearer} determines the syntactic role of each word in the sentence.

The GA runs for 1000 successive generations. In each one, the speaker samples three random words from the lexicon with their syntactic roles. Then, he samples an order according to the probability distribution in the grammar, concatenates the sampled words according to it, and sends it to the hearer. Then, a random noise flips every letter in the sentence with a probability of $p=0.01$. Note that the noisy sentence might contain words that are out of the lexicon.

Next, the hearer receives the noisy sentence and matches the syntactic roles in two steps. First, for every word in the sentence, he removes the noise by looking for the nearest word in the lexicon, in terms of Levenshtein distance. Then, he outputs the most probable word order given the grammar and the words.

Finally, we measure the Hamming distance between the input and the output by counting the correct syntactic roles. For example, the distance between SVO and SOV is $2/3$ since $O$, and $V$ are at different locations.

In each generation, 30\% of the grammars with the smallest distances are replicated to the next generation, with some random mutations to the probability distributions - a Gaussian noise with variance 0.01 is added to each probability in the grammar, and then re-normalized, to sum up to 1. After many generations, the surviving grammars are expected to be the ones that allow optimal communication between the speaker and the hearer.

We use this method to examine whether adding information to the sentence, such as the distinction between nouns and verbs or explicit case markers, changes the optimal grammar. 

We tested different values for the population size, selection rate, and mutation rate, but all the examined values yielded similar trends.

\section{Theoretical Analysis}

As mentioned, optimal communication occurs when the distance between the input and output is minimal.
In this section, we prove that under this definition of optimality, and assuming the hearer always identifies the words in the lexicon correctly despite the noise (as indeed happens with a high probability), the following holds

\begin{theorem}
Any optimal grammar has exactly one fixed word order.
\end{theorem}

\begin{proof}
Let $X_1 \dots X_n$ and $Y_1 \dots Y_n$ be random variables of the occurrence of possible word orders, associated with the speaker and the hearer, respectively. We define $$P(X_i=1)=P(Y_i=1)=p_i , \qquad   i=1,...,n.$$ Let $d_{ij}$ be the distance between $X_i$ and $Y_j$, defined as the number of syntactic roles with different positions between $X_i$ and $Y_j$.
Then, the expected distance is given by $$\mathbb{E}[d] = \sum_{(i, j)} d_{ij} P(X_i, Y_j),$$ where $P(X_i, Y_j)$ is the probability that the speaker chose order $X_i$ (with probability of $p_i$) and the hearer chose order $Y_j$ (with probability of $p_j$). Since the hearer has no additional information other than the observed sentence, he picks the most probable order from his grammar, independently of the speaker. Accordingly, $P(X_i,Y_j) = p_i p_j$, and therefore the expected value of the distance is given by $$\mathbb{E}[d] = \sum_{(i, j)} p_i d_{ij} p_j.$$ Note that ${p_id_{ij}p_j \geq 0,\forall i,j}$, since $p_i,p_j$ are probabilities, and $ d_{ij}$ is a distance. Since $d_{ij}=0$ only for $i=j$, to get the minimal expected distance $0$ it should hold $\forall i\ne j$ either $p_i = 0$ or $p_j = 0$. Therefore, $p_i$ can be greater than zero in exactly one entry. For other words, the only probability distributions that give an expected distance of zero have probability 1 for a particular word order $X_i$ and zero for the rest. Accordingly, the optimal grammar must have fixed word order.
\end{proof}

\section{Experiments}\label{sec:experiments}
We tested the model in four different scenarios.

\begin{enumerate}
    \item \textbf{Base} - all words are taken from a single lexicon, without distinction between nouns and verbs and without case markers.
    \item \textbf{N-V} - separate lexicons for nouns and verbs, which enables the hearer to distinguish between nouns and verbs, thus identifying the verb in the sentence.
    \item \textbf{Case} - each word in the sentence carries a suffix indicating its syntactic role (S, V or O).
    \item \textbf{N-V \& Case} - both 2 and 3, i.e., there is a distinction between nouns and verbs, and every word in the sentence carries a case marker.
\end{enumerate}

\begin{figure}
\centering
    \includegraphics[width=.95\linewidth]{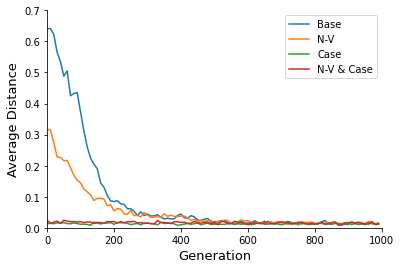}
\caption{Average distance of sentences over generations.}
\label{fig:dist}
\end{figure}

\section{Results}
For every scenario, we present the average distance between the input and the output among all grammars in each generation (Figure \ref{fig:dist}). It can be observed that the base scenario starts with the highest distance among the scenarios, as the sentences carry no information regarding the syntactic roles. A similar discussion could be held for the N-V scenario, in which the hearer can identify the verb in the sentence but cannot distinguish between the subject and object. In both scenarios, we observe a drop in the average distance, meaning that at some point, the hearers successfully parsed the sentences with only a small error.
In the two scenarios with the case markers, we see that the distance is very close to zero even in the first generation, meaning that the hearers parsed the sentences correctly from the first generation.  

\begin{figure}
\centering
\includegraphics[width=.95\linewidth]{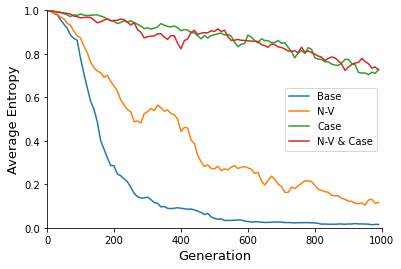}
\caption{Average entropy of grammars over generations.}
\label{fig:ent}
\end{figure}

To examine if the grammars converged to fixed word order, we present the grammars' average entropy in every generation (Figure \ref{fig:ent}). The entropy is defined by $H = -\sum_{i} p_i \log p_i$ and satisfies $0\leq H\leq 1$ where $H = 1$ represents a uniform distribution (absolute randomness) and $H = 0$ represents probability 1 for one word order and zero for the rest (absolute determinism). It can be observed that the average entropy of grammars begins at 1 (absolute randomness) in all scenarios, as the probability for every word order is equal. In the base scenario, we observe a fast decline in entropy until it converges to zero, i.e., a fixed word order. A decline is also observed in the N-V scenario. However, it is slower and does not converge to zero even after 1000 generations. As for the scenarios with the case markers, in both cases, we observe that the entropy remains high even after 1000 generations, meaning that no particular word order was fixed.

\begin{figure}
\centering
    \includegraphics[width=.95\linewidth]{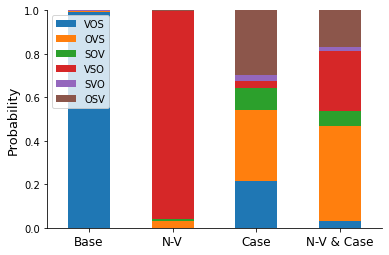}
    \caption{Word order probability after 1000 generations.}
\label{fig:ord}
\end{figure}

We also present the grammar that had the smallest distance after 1000 generations (Figure \ref{fig:ord}). In the base and N-V scenarios, we observe that the grammar converged to an (almost) fixed word order, VOS in the base scenario, and VSO in the N-V scenario. In contrast, in both scenarios with case markers, the word order remained flexible even after 1000 generations.

\section{Discussion}

In the base scenario, we observed a fast fixation of single word order. Therefore, in line with the theoretical result, we conclude that given no additional information regarding the sentence's syntactic relations, fixed word order is necessary to allow successful communication. We also observed a fixation in the N-V scenario, meaning that a distinction between nouns and verbs does not provide sufficient information for communication, and thus fixed word order is also necessary in this case.

In contrast, both scenarios with case markers did not converge to preferred word order, meaning that the case markers by themselves are enough to allow communication. This finding is consistent with the result of artificial language learning experiments conducted on human learners \cite{fedzechkina2011functional}. Note that since the case markers we used were just one letter, they could have been flipped by the random noise, leading to a wrong parsing by the hearer. However, the chances that two case markers in the same sentence will be flipped are small, and thus the hearer can parse the sentence correctly with high probability.

We note that additional experiments performed with a more significant number of syntactic roles yielded similar trends. Hence we conclude that these findings do not depend on the number of syntactic roles.

The reason why the base scenario converged to VOS and the N-V scenario converged to VSO is arbitrary, due to random drift, since there is no functional advantage to a particular word order over the others.

\section{Conclusions}
This work presents a simplistic functional model of language that demonstrates how a fixed word order in languages may result from a lack of information about syntactic relations.

Empirically, we used Genetic Algorithms (GA) in order to investigate whether additional information would render the fixed word order unnecessary. We showed that a distinction between nouns and verbs provides additional information to the hearer, but it is not sufficient. In contrast, case markers provide enough information to enable a flexible word order. These results are consistent with the typological findings.

Regarding the reason why most natural languages tend exhibit SVO or SOV orderings, the model used in this work does not seem to provide an explanation. It has been demonstrated that even where a single word order has been fixed, we have received various possible orders that are not necessarily one of these two. This is because we did not consider the functional advantages of certain orders over others, so the final order is entirely determined by random chance. Additional psycho-linguistic considerations are needed to explain the typological finding, as done in other studies \cite{gibson2013noisy, hengeveld2004parts}. 

\bibliographystyle{acl_natbib}
\bibliography{anthology}

\end{document}